\documentclass[10pt,twocolumn,twoside]{IEEEtran} 
\usepackage{times}
\usepackage{epsfig}
\usepackage{graphicx}
\graphicspath{{figures/}}
\usepackage{amsmath}
\usepackage{amssymb}

\usepackage{algorithm}
\usepackage{algorithmic}

\usepackage{multirow}
\usepackage{array}
\usepackage{rotating}

\usepackage{xcolor}
\usepackage{url}
\usepackage{float}
\usepackage{subfigure}

\usepackage[american]{babel}

\newcommand{\M}{{\mathcal{M}}}
\ifCLASSINFOpdf
\else
\fi

\begin{document}
%

\title{Enhance Visual Recognition under Adverse Conditions via Deep Networks}
%
%
%

\author{Ding~Liu,~\IEEEmembership{Member,~IEEE,}
        Bowen~Cheng,
        Zhangyang~Wang,~\IEEEmembership{Member,~IEEE,}
        Haichao~Zhang,~\IEEEmembership{Member,~IEEE,}
        and~Thomas~S.~Huang,~\IEEEmembership{Life~Fellow,~IEEE}
\thanks{The first two authors contributed equally to this work.}
\thanks{This work was supported in part by US Army Research Office grant W911NF-15-1-0317.}
\thanks{D. Liu, is with ByteDance Inc., Palo Alto,
CA, 94036 USA (e-mail: liudingdavy@gmail.com).}
\thanks{B. Cheng and T. S. Huang are with the Department
of Electrical and Computer Engineering and Beckman Institute, Univerisity of Illinois at Urbana-Champaign, Urbana,
IL, 61801 USA (e-mail: bcheng9@illinois.edu; t-huang1@illinois.edu).}
\thanks{Z. Wang is with the Department of Computer Science and Engineering, Texas A\&M University, TX 77843 USA (e-mail: atlaswang@tamu.edu).}
\thanks{H. Zhang is with Baidu Research, Sunnyvale, CA 94089 USA (e-mail: hczhang1@gmail.com).}
}

\maketitle

\begin{abstract}

Visual recognition under adverse conditions is a very important and challenging problem of high practical value, due to the ubiquitous existence of quality distortions during image acquisition, transmission, or storage.
While deep neural networks have been extensively exploited in the techniques of low-quality image restoration and high-quality image recognition tasks respectively, few studies have been done on the important problem of recognition from very low-quality images. 
This paper proposes a deep learning based framework for improving the performance of image and video recognition models under adverse conditions, using robust adverse pre-training or its aggressive variant. 
The robust adverse pre-training algorithms leverage the power of pre-training and generalizes conventional unsupervised pre-training and data augmentation methods. 
We further develop a transfer learning approach to cope with real-world datasets of unknown adverse conditions. 
The proposed framework is comprehensively evaluated on a number of image and video recognition benchmarks, and obtains significant performance improvements under various single or mixed adverse conditions. 
Our visualization and analysis further add to the explainability of results.

\end{abstract}


%
\IEEEpeerreviewmaketitle

\section{Introduction}
\label{sec:intro}

While the visual recognition research has made tremendous progress in recent years, most models are trained, applied, and evaluated on high-quality (HQ) visual data, such as the LFW \cite{LFW} and ImageNet \cite{Alex} benchmarks. 
However, in many emerging applications such as autonomous driving, intelligent video surveillance and robotics, the performances of visual sensing and analytics can be seriously endangered by different adverse conditions \cite{de2014face} in complex unconstrained scenarios, such as limited resolution, noise, occlusion and motion blur. 
For example, video surveillance systems have to rely on cameras of limited definitions, due to the prohibitive costs of installing high-definition cameras everywhere, leading to the practical need to recognize faces reliably from very low-resolution images \cite{VFR}. 
Other quality factors, such as occlusion and motion blur, are also known as critical concerns for commercial face recognition systems. 
As similar problems are ubiquitous for recognition tasks in the wild, it becomes highly desirable to investigate and improve the robustness of visual recognition systems to low-quality (LQ) image data.

\begin{figure}[t]
	\centering
	\begin{minipage}{0.47\textwidth}
		\centering {
			\includegraphics[width=\textwidth]{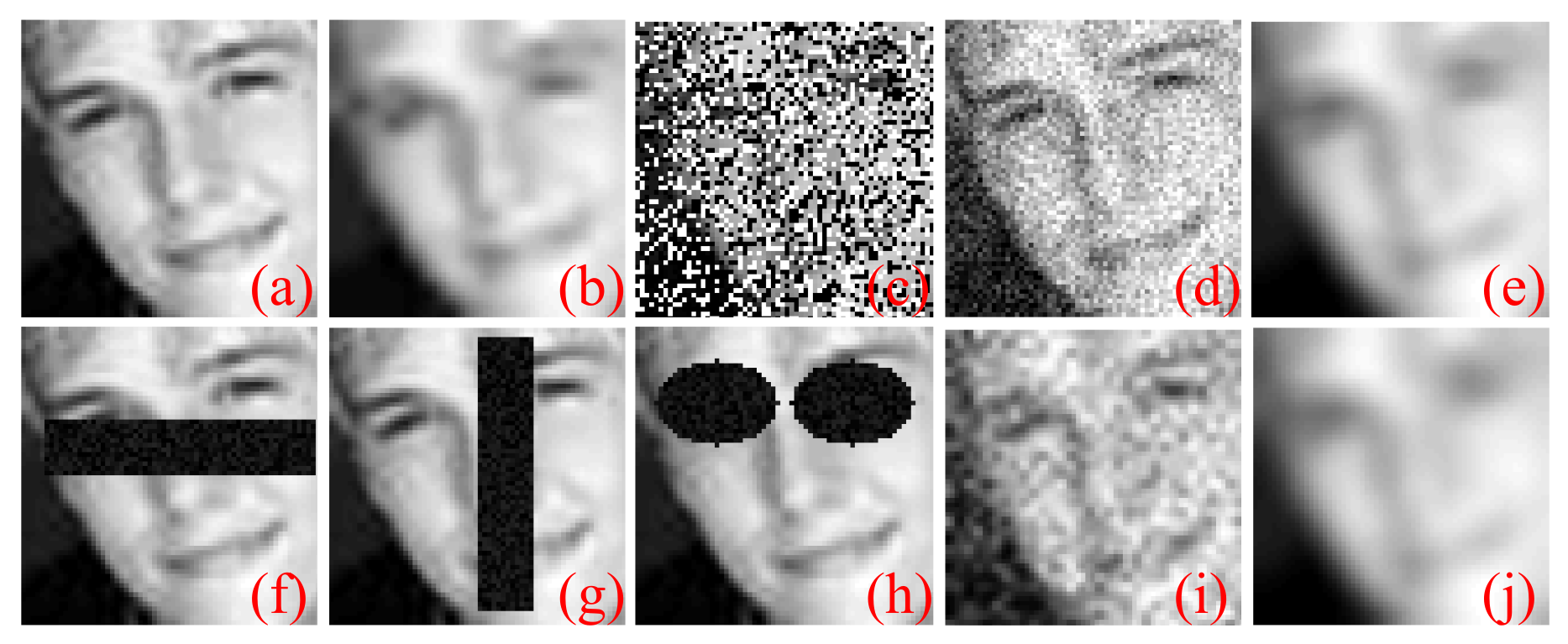}
	}\end{minipage}
	\caption{The original high-quality image from the MSRA-CFW dataset in (a), and (b) - (j) list various low-quality images generated from (a), that are all correctly recognized by our proposed models: (b) downsampled by a factor of 4; (c) 50\% salt \& pepper noise; (d) Gaussian noise (${\rm std} = 25$); (e) Gaussian blur (${\rm std} = 5$); (f)-(h) random synthetic occlusions; (i) downsampled by 4 followed by adding Gaussian noise (${\rm std} = 25$); (j) downsampled by 4 followed by adding Gaussian blur (${\rm std} = 5$).}
	\label{fig:intro}
\end{figure}

Unfortunately, exiting studies demonstrate that most state-of-the-art models appear fragile when applied on low-quality data.
The literature \cite{DVS12, ab14} has confirmed the significant effects of quality factors such as low-resolution, contrast, brightness, sharpness, focus, and illumination on commercial face recognition systems. 
The recent work \cite{karahan2016image} revealed that common degradations can even dramatically lower face recognition accuracy of the latest deep learning based face recognition models \cite{Alex, parkhi2015deep, szegedy2015going}. 
In particular, blur, noise, and 
occlusion cause the most significant performance deterioration.
Besides face recognition, the low-quality data is also found to adversely affect other recognition applications, such as hand-written digit recognition \cite{basu2015learning}, style recognition \cite{vlrr}, image classification \cite{liu2018image}.

This paper targets this important but less explored problem of visual recognition under adverse conditions. 
We study how and to what extent such adverse visual conditions can be coped with, aiming to improve the robustness of visual recognition systems on low-quality data. 
We carry out a comprehensive study on improving deep learning models for both image and video recognition tasks. 
We generalize conventional unsupervised pre-training and data augmentation methods, and propose the \textit{robust adverse pre-training} algorithms. 
The algorithms are generally applicable to various adverse conditions, and are jointly optimized with the target task. 
Figure~\ref{fig:intro}~(b)-(j) depict a series of heavily corrupted, low-quality images.  
They are all correctly recognized by our proposed models, though challenging even for human to recognize.

The major technical innovations are summarized in three aspects:
\begin{itemize}
	\item We present a framework for visual recognition under adverse conditions, that improves deep learning based models via robust pre-training and its aggressive variant. The framework is extensively evaluated on various datasets, settings and tasks. Our visualization and analysis further add to the explainability of results.
	\item We extend the framework to video recognition, and discuss how the temporal fusion strategy should be adjusted under different adverse conditions. 
	\item We develop a transfer learning approach for real-world datasets of unknown adverse conditions without synthetic LQ-HQ pairs directly available. We empirically demonstrates that our approach also improves the recognition on the original benchmark dataset. 
\end{itemize}


In the following, we will first review related work in Section~\ref{sec:related}.
Our proposed robust adverse pre-training algorithm and its variant, as well as the corresponding image based experiments are introduced in Section ~\ref{sec:image}.
Video based experiments are reported with implementation details in Section~\ref{sec:video}.
The transfer learning approach for dealing with real-world datasets is described in Section~\ref{sec:unknown}.
Finally, conclusions and discussions are provided in Section~\ref{sec:conc}.
\section{Related Work}
\label{sec:related}

\subsection{Visual Recognition under Adverse Conditions}

In a real-world visual recognition problem, there is indeed no absolute boundary between LQ and HQ images. Yet as commonly observed, while some mild degradations may have negligible impact on the recognition performance, the impact turns much notable once the level of adverse conditions passes some empirical threshold. The object and scene recognition literature reports a significant performance drop when the image resolution is decreased below $32 \times 32$ pixels \cite{80tiny}. In \cite{VFR}, the authors find the face recognition performance to be notably deteriorated when face regions become smaller than $16 \times 16$ pixels. \cite{karahan2016image} reports a rapid decline of face recognition accuracies, with Gaussian noise of standard deviation (std) between $10$ and $20$. \cite{DVS12, ab14} reveal more impacts of contrast, brightness, sharpness, and out-of-focus on image based face recognition. 

To resolve that, the conventional approach first resorts to image restoration and then feeds the restored image into a classifier \cite{fergus2006removing,yang2010image,liu2017robust}. Such a straightforward approach yields the sub-optimal performance: the artifacts introduced by the reconstruction process will undermine the final recognition. \cite{VFR, cvpr08} incorporate class-specific features in the restoration as a prior. 
%
A domain adaptation scheme of pre-training is utilized for sentiment classification \cite{glorot2011domain} and font recognition \cite{wang2015deepfont}, where a sub-model is first trained in a unsupervised manner from both LQ and HQ data, aiming to minimize the low-level feature difference between them.
%
\cite{zhang2011close} presents a joint image restoration and recognition method, based on the assumption that the degraded images, if correctly restored, also have a good identifiability.
A similar approach is adopted for jointly dealing with image dehazing and object detection in \cite{li2017aod}.
Those ``close-the-loop'' ideas achieve superior performance over the traditional two-stage pipelines. 

Compared to single image object recognition, the impact of adverse conditions on video recognition is as profound and significant, with many attentions paid to tasks such as video face recognition and tracking \cite{stasiak2009face}, license plate recognition \cite{chen2007license}, and facial expression recognition \cite{tian2004evaluation}. \cite{shan2005recognizing} introduces robust hand-crafted features to low-resolution and head motion blur. \cite{arandjelovic2007manifold} combines a shape-illumination manifold framework with implicit super-resolution. \cite{herrmann2016low} adopts a residual neural network trained with synthetic LQ samples, which are generated by a controlled corruption process such as adding motion blur or compression artifacts.

\subsection{Deep Networks under Adverse Conditions}

Convolutional neural networks (CNNs) have gained explosive popularity in recent years for visual recognition tasks \cite{Alex, karpathy2014large}. However, their robustness to adverse conditions remain unsatisfactory \cite{karahan2016image}. Deep networks are shown to be susceptible to adversarial samples \cite{goodfellow2014explaining}, generated by introducing carefully chosen perturbations to the input. 
Besides that,
the common  adverse conditions, stemming from artifacts during image acquisition, transmission, or storage, still easily mislead deep networks in practice \cite{dodge2016understanding}. \cite{karahan2016image} confirms the fragility of the state-of-the-art deep face recognition models \cite{Alex, parkhi2015deep, szegedy2015going}, to various adverse conditions, in particular blur, noise, and periocular region occlusion. 
Besides face recognition, the adverse conditions are also found to negatively affect other recognition tasks, such as hand-written digit recognition \cite{basu2015learning} and style recognition \cite{vlrr}. 

While data augmentation has become a standard tool \cite{Alex}, the primary goal is to artificially  increase the training data volume and improve the model generalization. The augmentation methods are moderate in practice, by adding small noise or pixel translations, etc. The learned model is then to be applied on clean HQ images for testing. Those methods are thus not dedicated to handling specific types of severe degradation.
%
Unsupervised pre-training \cite{vincent2008extracting,erhan2009difficulty} also effectively regularizes the training process, especially when labeled data is insufficient. Classical pre-training methods reconstruct the input data from itself \cite{erhan2009difficulty} or its slightly transformed versions \cite{vincent2008extracting,masci2011stacked} from auto-encoders. 
However, these methods are focused more on improving model generalization on standard (HQ) data, but not the LQ data that suffers a certain type of severe degradation. 
%
The recent work \cite{vlrr} describes an approach of
pre-training a deep network model for image recognition under the low-resolution case. 
However, it  neither considers any other type of adverse conditions or mixed degradations,\footnote{The  solutions to low-resolution cases cannot be straightforwardly extended to other adverse conditions. For example, we tried Model III of \cite{vlrr} in salt \& pepper noise and occlusion cases, finding the performance to be hurt sometimes.} nor takes into account any video based problem setting. Most crucially, \cite{vlrr} required pairs of synthetic training samples before and after degradation. While the degradation process is unknown in real-world data, the applicability of the proposed algorithm is severely limited.

\section{Image Based Visual Recognition under Single or Mixed Adverse Conditions} 
\label{sec:image}

\subsection{Problem Statement}

We start by introducing single image based visual recognition models in this section, and extend to the video recognition models later. 
We define the visual recognition model $\M$ that predicts the category labels $\{l_i\}_{i=1}^N$ from the images $\{\mathbf{y}_i\}_{i=1}^N$. Due to the adverse conditions, $\{\mathbf{y}_i\}_{i=1}^N$ can be viewed as low-quality (LQ) images, degraded from high-quality (HQ) ground truth images $\{\mathbf{x}_i\}_{i=1}^N$. For now, we treat the original training datasets as HQ images $\{\mathbf{x}_i\}$, and generate LQ images $\{\mathbf{y}_i\}$ using synthetic degradation. In testing, our model operates with only LQ inputs.

We define a CNN based image recognition model $\M$ with $d$ layers. The first $d_1$ layers are convolutional, while the remaining $d - d_1$ layers are fully connected. The $i$-th convolutional layer, denoted as $conv_i$ ($i = 1, \cdots, d_1$), contains $n_i$ filters of size $c_i \times c_i$, with default stride size 1 and zero-padding. The $j$-th fully connected (fc) layer, denoted as $fc_j$ ($j =  1, \cdots, d - d_1$), has $m_j$ nodes. We use ReLU activation and apply dropout with a rate of 0.5 to fully connected layers. Cross-entropy loss is adopted for classification, while mean square error (MSE) is used for reconstruction.


\subsection{Robust Adverse Pre-training of Sub-models}

Building a classifier $\M$ directly on $\{\mathbf{y}_i\}$ is usually not robust due to the severe information loss  caused by adverse conditions. Training $\M$ over \{$\{\mathbf{x}_i\}$, $\{l_i$\}\} also does not perform well when tested on $\{\mathbf{y}_i\}$ due to the domain mismatch \cite{vlrr, VFR}. Our main intuition is to regularize and enhance the feature extraction from $\{\mathbf{y}_i\}$, via injecting auxiliary information from $\{\mathbf{x}_i\}$. With the help of $\{\mathbf{x}_i\}$, the model better discriminates the true signal from the severe corruption, and learns more robust filters from low-quality inputs. The entire $\M$ can be well adapted for the mapping from $\{\mathbf{y}_i\}$ to $\{l_i$\} by a joint optimization step followed.

To pre-train $\M$, we first define the sub-model $\M_s$ with $k$ layers. Its first $k_p$ layers are configured the same as the first $k_p$ layers from $\M$. The last $k - k_p$ layers reconstruct the input image from the output feature maps of the $k_p$-th layer. 
We generate $\{\mathbf{y}_i\}$ from $\{\mathbf{x}_i\}$, based on a degradation process parameterized by the \textit{adverse factor} $\alpha$\footnote{Here the \textit{adverse factor} is defined in a broad sense. It can be the downsampling factor for low-resolution, the proportion of image for noise corruption, the degree of blur and so on.}, in order to meet the adverse conditions in testing. 
We then train $\M_s$ to reconstruct $\{\mathbf{x}_i\}$ from $\{\mathbf{y}_i\}$. 
We empirically find that pre-training only a part of convolutional layers (i.e., $k_p \leq d_1$) maintains a good balance between the feature extraction and the discrimination ability, with the best performance. After $\M_s$ is trained, its first $k_p$ layers are exported to initialize the first $k_p$ layers of $\M$. $\M$ is then jointly tuned for the recognition task over $\{\mathbf{y}_i, \{l_i\}$. The algorithm, termed as \textit{Robust Adverse Pre-training} (\textbf{RAP}), is outlined in Algorithm \ref{pretrain}.

\begin{algorithm}[h]
	\caption{Robust adverse pre-training}
	\label{pretrain}
	\begin{algorithmic}[1]
		\REQUIRE Configuration of $\M$; $\{\mathbf{x}_i\}$ and $\{l_i\}$, $i$ = $1, 2,..., N$; the choice of $k$; the adverse factor $\alpha$.
		\STATE Generate $\{\mathbf{y}_i\}$ from $\{\mathbf{x}_i\}$, based on a degradation process parameterized by $\alpha$
		
		\STATE Construct the $k$-layer sub-model $\M_s$. Its first $k_p$ layers are configured identically to those of $\M$.
		
		\STATE Train $\M_s$ to reconstruct $\{\mathbf{x}_i\}$ from $\{\mathbf{y}_i\}$, under MSE.
		
		\STATE Export the first $k_p$ layers from $\M_s$ to initialize  the first $k_p$ layers of $\M$, where $k_p < k$.
		
		\STATE Tune $\M$ over \{$\{\mathbf{y}_i\}$, $\{l_i$\}\}, under the cross-entropy loss.
		
		\ENSURE $\M$.
	\end{algorithmic}
\end{algorithm}

\subsection{Aggressively Robust Adverse Pre-training}

Different from testing when only LQ data is available, we have the flexibility to synthesize LQ images for training at our will. 
While the RAP algorithm  trains $\M$ and $\M_s$ under the same adverse condition, 
we continue to explore when the $\M_s$ pre-training and $\M$ joint-tuning are performed under different levels of adverse conditions. 
This is motivated by the denoising autoencoders~\cite{vincent2010stacked}, where the pre-training was conducted by noisy data and the subsequent classification model was learned with clean data. Our conjecture is that pre-training $\M_s$ in severer degradation can actually help $\M_s$ capture more robust feature mappings. 
This leads to the \textit{Aggressively Robust Adverse Pre-training} (\textbf{ARAP}), a variant of RAP, outlined in Algorithm \ref{dpretrain}. We assume the degradation process of $\{\mathbf{y}_i\}$ to be identical to the target testing data, while $\{\mathbf{z}_i\}$ is a more heavily degraded set independently generated from $\{\mathbf{x}_i\}$. 
The larger adverse factor indicates the severer degradation, 
and thus in this case the adverse factor $\beta$ for generating $\{\mathbf{z}_i\}$ is larger than $\alpha$ for $\{\mathbf{y}_i\}$. 
RAP can be a special case of ARAP where $\alpha$ and $\beta$ coincide. 


\begin{algorithm}[ht]
	\caption{Aggressively robust adverse pre-training}
	\label{dpretrain}
	\begin{algorithmic}[1]
		\REQUIRE Configuration of $\M$; $\{\mathbf{x}_i\}$ and $\{l_i\}$, $i$ = $1, ..., N$; the choice of $k$; two adverse factors $\alpha$ and $\beta$ ($\beta > \alpha$).
		
		\STATE Generate $\{\mathbf{y}_i\}$, $\{\mathbf{z}_i\}$ from $\{\mathbf{x}_i\}$, based on two degradation processes parameterized by $\alpha$ and $\beta$, respectively.
		
		\STATE Construct the sub-model $\M_s$ same as in Algorithm 1.
		
		\STATE Train $\M_s$ to reconstruct $\{\mathbf{x}_i\}$ from $\{\mathbf{z}_i\}$, under MSE.
		
		\STATE Export the first $k_p$ layers from $\M_s$ to initialize the first $k_p$ layers of $\M$, where $k_p < k$.
		
		\STATE Tune $\M$ over \{$\{\mathbf{y}_i\}$, $\{l_i$\}\}, under the cross-entropy loss.
		
		\ENSURE $\M$.
	\end{algorithmic}
\end{algorithm}

\subsection{Experiments on Benchmarks}

\subsubsection{Object Recognition on the CIFAR-10 Dataset}

\begin{table*}
	\fontsize{10pt}{12pt}\selectfont
	\caption{The top-1 and top-5 classification accuracy (\%) on the CIFAR-10 dataset, where LQ images are generated by downsampling the original images with a factor of $\alpha$ = 2.}
	\begin{center}
		\begin{tabular}{c|c|c|c|c|c|c|c|c}
			\hline
			& HQ & LQ-2 & RAP-2-non-joint & RAP-2 & ARAP-2-4 & ARAP-2-8 & ARAP-2-12 & ARAP-2-16 \\ 
			\hline
			\hline
			Top-1  & 67.43 & 60.79 & 46.89 & 62.12 & 62.80 & \textbf{63.31} & 62.91 & 62.56 \\
			Top-5  & 96.61 & 95.32 & 90.77 & 95.10 & 95.52 & \textbf{95.80} & 95.34 & 95.10 \\ 
			\hline
		\end{tabular}
	\end{center}
	
	\label{CIFARlr2}
\end{table*}

\begin{table}
	\fontsize{10pt}{12pt}\selectfont
	\caption{The top-1 and top-5 classification accuracy (\%) on the CIFAR-10 dataset, where LQ images are generated by adding $\alpha$ = 50\% salt \& pepper noise.}
	\begin{center}
		\begin{tabular}{@{\hskip 1mm}c@{\hskip 1mm}|@{\hskip 1mm}c@{\hskip 1mm}|@{\hskip 1mm}c@{\hskip 1mm}|@{\hskip 1mm}c@{\hskip 1mm}|@{\hskip 1mm}c@{\hskip 1mm}}
			\hline
			& HQ & LQ-50\% & RAP-50\%-non-joint & RAP-50\% \\ \hline
			\hline
			Top-1  & 67.43 & 33.46 & 38.64 & \textbf{50.32} \\ 
			Top-5  & 96.61 & 83.22 & 86.86 & \textbf{92.03} \\ \hline
		\end{tabular}
	\end{center}
	
	\label{CIFARsalt}
\end{table}

In order to validate our algorithm, we first conduct object recognition on
the CIFAR-10 dataset \cite{cifar}, which consists of  60,000 color images of $32 \times 32$ pixels from 10 classes (we convert all to grayscale ones). Each class has 5,000 training images and 1,000 test images. We generate LQ images as per each specific type of adverse conditions, where the adverse factors $\alpha$ or $\beta$ become concrete degradation hyper-parameters such as downsampling factor, noise level, or blur kernel. We perform no other data augmentation beyond generating LQ images. 

We choose $\M$ with $d = 4$, with $d_1 = 3$ convolutional layers, followed by $d - d_1 = 1$ fully connected layer with $m_1$ always equaling the number of classes. 
Unless otherwise stated, we set $\M_s$ as a fully convolutional network with the empirical values $k=3, k_p = 2$, which work well in all experiments. The default configuration of convolutional layers are: $n_1$ = 64, $c_1$ = 9; $n_2$ = 32, $c_2$ = 5; $n_3$ = 20, $c_3$ = 5. We first train $\M_s$ with learning rate 0.0001, and then jointly tune $\M$ with a learning rate 0.001 for the first $k_p$ layers and 0.01 for the rest $d - k_p$ layers. Both learning rates are reduced by a factor of 10 every 5,000 iterations. 

\paragraph{Low-Resolution}
We generate LQ (low-resolution) images $\{\mathbf{y}_i\}$ by following the process in \cite{dong2014learning,liu2016robust}: first downsampling the HQ (high-resolution) images $\{\mathbf{x}_i\}$ by a factor of $\alpha$, then upsampling back to the original size with bicubic interpolation. 
We use the same process for all the following experiments of low-resolution degradation, unless otherwise stated.
We compare the following approaches:
\begin{itemize}
	\item \textbf{HQ:} $\M$ is trained and tested on $\{\{\mathbf{x}_i\}, \{l_i\}\}$. 
	\item \textbf{LQ-$\alpha$:} $\M$ is trained and tested on $\{\{\mathbf{y}_i\}, \{l_i\}\}$. 
	\item \textbf{RAP-$\alpha$-non-joint:} $\M_s$ is pre-trained using the Step 3 of Algorithm \ref{pretrain} on $\{\{\mathbf{y}_i\},\{\mathbf{x}_i\}\}$. The remaining $d-k_p$ layers of $\M$ are then trained on $\{\{\mathbf{y}_i\}, \{l_i\}\}$, with the first $k_p$ pre-trained layers fixed. It is identical to RAP except for no jointly tuning $\M$. 
	\item \textbf{RAP-$\alpha$:}  $\M$ is trained using RAP (Algorithm \ref{pretrain}).
	\item \textbf{ARAP-$\alpha$-$\beta$:}  $\M$ is trained using ARAP (Algorithm \ref{dpretrain}), where $\beta$ is a larger downsamping factor than $\alpha$.
\end{itemize}
The evaluation of $\M$s is all performed on the testing set of LQ images (except for the HQ baseline), downsampled by the factor $\alpha$. The first two baselines aim to examine how much the adverse condition affects the performance. 

Table \ref{CIFARlr2} displays the results at $\alpha = 2$, which is a challenging problem of recognizing objects from   images of $16\times16$ pixels. Such an adverse condition dramatically affects the performance, by dropping the top-1 accuracy for nearly 7\%, after comparing LQ-2 with HQ. It might be unexpected that the performance of RAP-2-non-joint is much inferior to that of LQ-2.  As observed in this and many following experiments, the reconstruction based pre-training step, if not jointly optimized for the recognition step, often hurts the performance rather than does any help. By adding the joint tuning step, RAP-2 gains a 1.33\% advantage over LQ-2 in the top-1 accuracy, which is owning to the $\M_s$ pre-training that involves auxiliary yet beneficial information from HQ data. 

It is noteworthy that all four ARAP methods ($\beta = 4, 8, 12, 16$) show superior results over RAP-2. ARAP-2-8 achieves the best accuracy of 63.31\% (top-1) and 95.80\% (top-5). The observation confirms our conjecture that more robust feature extractions could be achieved by purposely pre-training $\M_s$ in severer degradation ($\beta > \alpha$). As $\beta$ grows with $\alpha$ fixed at 2, the performance of ARAP first improves and then drops, with the peak at $\beta = 8$. That is also explainable, since if $\{\mathbf{z}_i\}$ are too much degraded, little information is left for training $\M_s$. 


\begin{table}[t]
	\fontsize{10pt}{12pt}\selectfont
	\caption{The top-1 and top-5 face identification accuracy (\%) on the MSRA-CFW dataset, where LQ images are generated by adding $\alpha$ = 50\% salt \& pepper noise.}
    \centering
		\begin{tabular}{@{\hskip 1mm}c@{\hskip 1mm}|@{\hskip 1mm}c@{\hskip 1mm}|@{\hskip 1mm}c@{\hskip 1mm}|@{\hskip 1mm}c@{\hskip 1mm}|@{\hskip 1mm}c@{\hskip 1mm}|@{\hskip 1mm}c@{\hskip 1mm}}
			\hline
			\# of pre-trained layers & 0 & 1 & 2 & 3 & 4 \\ \hline
			\hline
			Top-1  & 14.75 & 27.04 & \textbf{49.86} & 42.17 & 38.64 \\ 
            
			Top-5  & 36.28 & 45.36 & \textbf{72.14} & 63.09 & 59.85 \\ \hline
		\end{tabular}
	
	\label{tab:no_pretrain}
\end{table}

\begin{table*}
	\fontsize{10pt}{12pt}\selectfont
	\caption{The top-1 and top-5 classification accuracy (\%) on the CIFAR-10 dataset, where LQ images are generated by blurring original images (HQ), with Gaussian kernel of std $\alpha$ = 2.}
	\begin{center}
		\begin{tabular}{c|c|c|c|c|c|c|c}
			\hline
			& HQ & LQ-2 & RAP-2-non-joint & RAP-2 & RAP-2-5 & RAP-2-8 & RAP-2-9 \\ \hline
			\hline
			Top-1  & 67.43 & 52.62 & 39.80 & 54.73 & 54.77 & \textbf{55.67} & 54.35 \\ 
			Top-5  & 96.61 & 92.70 & 87.34 & 93.24 & 93.50 & \textbf{93.52} & 93.15 \\ 
			\hline
		\end{tabular}
	\end{center}
	
	\label{CIFARblur2}
\end{table*}

\paragraph{Noise}
Since adding moderate Gaussian noise has been standard for data augmentation, we focus on the more destructive salt \& pepper noise. The LQ images $\{\mathbf{y}_i\}$ are generated by randomly choosing $\alpha = 50\%$ pixels in each HQ image $\mathbf{x}_i$ to be replaced with either 0 or 255. We compare HQ, LQ-$\alpha$, RAP-$\alpha$-non-joint, and RAP-$\alpha$, all of which are similarly defined as in the low-resolution case. We tried RAP-$\alpha$-$\beta$, but did not get much performance improvement over RAP-$\alpha$ as we did for low-resolution. In Table \ref{CIFARsalt}, the severe information loss by 50\% salt \& pepper noise is reflected on the 34\% top-1 accuracy drop from HQ to LQ-50\%. After only pre-training the first few layers, there is a 5.18\% increase in the top-1 accuracy, obtained by RAP-50\%-non-joint. RAP-50\% achieves the closest accuracy to the HQ baseline, and outperforms RAP-50\%-non-joint by 11.68\% and 5.17\%, in terms of top-1 and top-5 accuracy, respectively. Those results re-confirm the necessity of both per-training and end-to-end tuning for RAP.

To determine the number of pre-trained layers in $\M$, we conduct an ablation study by changing the number of pre-trained layers in the model under the adverse condition of salt \& pepper noise. The top-1 and top-5 face identification accuracy (\%) on the MSRA-CFW dataset is shown in Table~\ref{tab:no_pretrain}. It is observed that when there are very few layers pre-trained, the model extracts very limited information from the LQ-HQ reconstruction. In contrast, when too many layers are pre-trained, the model is very biased to the reconstruction task and less suitable to the recognition task. We choose pre-training the first two layers in our experiments, which achieves the highest top-1 and top-5 accuracy.

\paragraph{Blur}
Images commonly suffer from various types of blurs, such as simple Gaussian blur, motion blur, out-of-focus blur, or their complex combinations \cite{zhang2011close}. We focus on the Gaussian blur, while similar strategies can be naturally extended to other types. The LQ images $\{\mathbf{y}_i\}$ are generated by convolving the HR images $\{\mathbf{x}_i\}$ with a Gaussian kernel with std $\alpha = 2$, and the fixed kernel size of $9\times9$ pixels. We compare HQ, LQ-$\alpha$, RAP-$\alpha$-non-joint, RAP-$\alpha$, and ARAP-$\alpha$-$\beta$ ($\beta$ denotes a larger std than $\alpha$), all similarly defined. 

Table \ref{CIFARblur2} demonstrates similar findings as the low-resolution case. The non-adapted restoration in RAP-$\alpha$-non-joint only leaves it worse than LQ-$\alpha$. RAP-$\alpha$ gains 1.21\% over LQ-$\alpha$ in top-1 accuracy. Two out of three ARAP methods ($\beta = 5, 8$) yield greatly improved results than RAP-$\alpha$, while $\beta = 9$ is only marginally inferior. Using Algorithm \ref{dpretrain}, $\M_s$ trained with heavier blurs 
tends to produce more discriminative features, when applied to LQ data with lighter blurs, which benefits recognition tasks.

%

\begin{table}
	\caption{The top-1 and top-5 face identification accuracy (\%) on the MSRA-CFW dataset, where LQ images are generated by downsampling original images by a factor of $\alpha$ = 4.}
    \resizebox{\columnwidth}{!}{    
		\begin{tabular}{@{\hskip 1mm}c@{\hskip 1mm}|@{\hskip 1mm}c@{\hskip 1mm}|@{\hskip 1mm}c@{\hskip 1mm}|@{\hskip 1mm}c@{\hskip 1mm}|@{\hskip 1mm}c@{\hskip 1mm}|@{\hskip 1mm}c@{\hskip 1mm}|@{\hskip 1mm}c@{\hskip 1mm}}
			\hline
			& HQ & LQ-4 & CDP & RAP-4-non-joint & RAP-4 & ARAP-4-6\\ \hline
			\hline
			Top-1  & 57.25 & 50.79 & 53.61 & 50.50 & \textbf{54.23} & 54.10\\ 
			Top-5  & 76.89 & 72.81 & 73.82 & 72.88 & 74.06 & \textbf{74.97}\\ \hline
		\end{tabular}
	}
	
	\label{msralr}
\end{table}

\begin{table}
	\caption{The top-1 and top-5 face identification accuracy (\%) on the MSRA-CFW dataset, where LQ images are generated by adding $\alpha$ = 50\% salt \& pepper noise.}
    \resizebox{\columnwidth}{!}{
		\begin{tabular}{@{\hskip 1mm}c@{\hskip 1mm}|@{\hskip 1mm}c@{\hskip 1mm}|@{\hskip 1mm}c@{\hskip 1mm}|@{\hskip 1mm}c@{\hskip 1mm}|@{\hskip 1mm}c@{\hskip 1mm}|@{\hskip 1mm}c@{\hskip 1mm}}
			\hline
			& HQ & LQ-50\% & CDP & RAP-50\%-non-joint & RAP-50\% \\ \hline
			\hline
			Top-1  & 57.25 & 14.75 & 17.54 & 26.20 & \textbf{49.86} \\ 
			Top-5  & 76.89 & 36.28 & 40.38 & 51.59 & \textbf{72.14} \\ \hline
		\end{tabular}
      }
	
	\label{msrasalt}
\end{table}

\begin{table}
	\caption{The top-1 and top-5 face identification accuracy (\%) on the MSRA-CFW dataset, where LQ images are generated by blurring the original images (HQ), with Gaussian kernel of std $\alpha$ = 5.}
    \resizebox{\columnwidth}{!}{
		\begin{tabular}{@{\hskip 1mm}c@{\hskip 1mm}|@{\hskip 1mm}c@{\hskip 1mm}|@{\hskip 1mm}c@{\hskip 1mm}|@{\hskip 1mm}c@{\hskip 1mm}|@{\hskip 1mm}c@{\hskip 1mm}|@{\hskip 1mm}c@{\hskip 1mm}|@{\hskip 1mm}c@{\hskip 1mm}}
			\hline
			& HQ & LQ-5 & CDP & RAP-5-non-joint & RAP-5 & ARAP-5-8 \\ \hline
			\hline
			Top-1  & 57.25 & 49.96 & 51.04 & 45.66 & \textbf{52.19} & 51.94\\ 
			Top-5  & 76.89 & 72.51 & 72.46 & 69.08 & 73.73 & \textbf{73.88}\\ \hline
		\end{tabular}
    }
	
	\label{msrablur}
\end{table}

\begin{table}
	\fontsize{10pt}{12pt}\selectfont
	\caption{The top-1 and top-5 accuracy (\%) on MSRA-CFW, where LQ images are generated with random synthetic occlusions.}
	\begin{center}
		\begin{tabular}{c|c|c|c|c}
			\hline
			& HQ & LQ-$\alpha$ & RAP-$\alpha$-non-joint & RAP-$\alpha$ \\ \hline
			\hline
			Top-1  & 59.41 & 32.62 & 34.91 & \textbf{43.96} \\ 
			Top-5  & 78.11 & 56.32 & 60.16 & \textbf{67.20} \\ \hline
		\end{tabular}
	\end{center}
	
	\label{msrahole}
\end{table}

\subsubsection{Face Identification on the MSRA-CFW Dataset}


We conduct face identification on the MSRA Dataset of Celebrity Faces on the Web (MSRA-CFW) \cite{zhang2012finding}, 
which includes $202, 792$ cropped and centered face images of $64\times64$ pixels in around $1600$ classes. We select a subset including all the 123 classes of more than $300$ images, to ensure the sufficient amount of training data for our deep network model. We split 90\% images of each class for training and 10\% for testing. We perform the face identification task, under highly challenging adverse conditions, such as very low resolution, noise, blur, occlusion and mixed cases. The visual examples are displayed in Figure~\ref{fig:intro}.

For the low resolution, noise or blur case, we set $\M$ with $d = 8$,  $d_1 = 6$. The convolutional layers are configured as: $n_1 = 32$, $c_1 = 9$;  $n_2 = 16$, $c_2 = 5$;  $n_3 = 20$, $c_3 = 4$; $n_4 = 40$, $c_4 = 3$; $n_5 = 60$, $c_5 = 3$; $n_6 = 80$, $c_6 = 2$. $fc_1$ has $m_1 = 160$ and $fc_2$ has $m_2 = 123$. For occlusion, we modify $n_1 = 16$, $c_1 = 21$;  $n_2 = 8$, $c_2 = 1$, and leave other six layers unchanged. Here the low-level filters perform in-painting, and thus needs larger receptive fields to predict missing pixels from neighborhoods. 


\paragraph{Low-Resolution, Noise, Blur}
The three adverse conditions follow similar settings and comparison methods to CIFAR-10. 
To validate the effectiveness of RAP and ARAP, we utilize for pre-training the domain adaptation scheme in \cite{glorot2011domain,wang2015deepfont}, termed Cross-Domain Pre-training (CDP), as an additional baseline method for comparison.
We adopt a larger downsampling factor 4 in the low-resolution case, and a larger blur std 5 for the blur case. The conclusions drawn from Tables \ref{msralr}, \ref{msrasalt} and \ref{msrablur} are also consistent to those of CIFAR-10: RAP boosts performance in all cases compared to LQ, CDP, and RAP-non-joint, and ARAP achieves considerably higher results for the two cases of low-resolution and blur.

%
%
%

\paragraph{Occlusion}
Prior studies in \cite{karahan2016image} discovered that the periocular occlusion degraded the face recognition performance most. We follow \cite{karahan2016image} to synthesize the occlusions for the periocular regions, in the shape of either rectangle or ellipse (chosen with equal probability). The size of either shape, as well as the pixel values within the synthetic occlusion, is drawn from uniform distributions. The center locations of synthetic occlusions are picked randomly in a bounding box, whose boundaries are determined by eye landmark points. We emphasize that the occlusion masks are unknown and changing for both training and testing, corresponding to the toughest blind inpainting problem \cite{xie2012image}. 

We evaluate HQ, LQ-$\alpha$, RAP-$\alpha$-non-joint and RAP-$\alpha$ in Table \ref{msrahole}. The parameter $\alpha$ generally denotes the controlled shape/size/location variations. We also tried large $\beta$ via enlarging the maximal size of occlusions, but observed no visible improvement from ARAP-$\alpha$-$\beta$. The occlusion causes much worse corruptions than previous adverse conditions: it completely masks a facial region that is known to be critical for recognition. The lost pixel information is harder to be restored than the salt \& pepper noise case, due to the missing neighborhood. As expected, the challenging random occlusions result in very significant drops from HQ to LQ. RAP-non-joint only marginally raises the accuracy (e.g., 2\% in top-1). RAP achieves the most encouraging improvements of 11.34\% and 10.88\%, in terms of top-1 and top-5 accuracy, respectively.

\begin{table}
	\fontsize{10pt}{12pt}\selectfont
	\caption{The top-1 and top-5 accuracy (\%) on MSRA-CFW, where LQ images are generated by first downsampling original images by $\alpha$ = 2 and then adding Gaussian noise with std 25.}
	\begin{center}
		\begin{tabular}{@{\hskip 1mm}c@{\hskip 1mm}|@{\hskip 1mm}c@{\hskip 1mm}|@{\hskip 1mm}c@{\hskip 1mm}|@{\hskip 1mm}c@{\hskip 1mm}|@{\hskip 1mm}c@{\hskip 1mm}|@{\hskip 1mm}c@{\hskip 1mm}}
			\hline
			& HQ & LQ-2 & RAP-2-non-joint & RAP-2 & ARAP-2-4 \\ \hline
			\hline
			Top-1  & 57.25 & 45.57 & 44.30 & 48.63 & \textbf{50.34}\\ 
			Top-5  & 76.89 & 69.82 & 68.00 & 71.89 & \textbf{73.76}\\ \hline
		\end{tabular}
	\end{center}
	
	\label{msragausslr}
\end{table}

\begin{table}
	\fontsize{10pt}{12pt}\selectfont
	\caption{The top-1 and top-5 accuracy (\%) on MSRA-CFW, where LQ images are generated by first downsampling original images by $\alpha$ = 4 and then blurred with the Gaussian kernel of std 2.}
	\begin{center}
		\begin{tabular}{@{\hskip 1mm}c@{\hskip 1mm}|@{\hskip 1mm}c@{\hskip 1mm}|@{\hskip 1mm}c@{\hskip 1mm}|@{\hskip 1mm}c@{\hskip 1mm}|@{\hskip 1mm}c@{\hskip 1mm}|@{\hskip 1mm}c@{\hskip 1mm}}
			\hline
			& HQ & LQ-4 & RAP-4-non-joint & RAP-4 & ARAP-4-8\\ \hline
			\hline
			Top-1  & 57.25 & 49.39 & 48.76 & 52.30 & \textbf{52.68}\\ 
			Top-5  & 76.89 & 71.29 & 70.99 & 73.80 & \textbf{74.51}\\ \hline
		\end{tabular}
	\end{center}
	
	\label{msrablurlr}
\end{table}

\begin{table*}[!h]
	\fontsize{10pt}{12pt}\selectfont
	\caption{The top-1 and top-5 digit recognition accuracy (\%) on the SVHN dataset, where LQ images are generated by blurring the original images (HQ), with the Gaussian kernel of standard deviation $\alpha$ = 2.}	
	\begin{center}
		\begin{tabular}{c|c|c|c|c|c|c}
			\hline
			& HQ & LQ-2 & RAP-2-non-joint & RAP-2 & ARAP-2-5 & ARAP-2-8 \\ 
			\hline
			\hline
			Top-1 & 89.23 & 85.40 & 83.84 & 82.47 & \textbf{89.40} & 88.29 \\
			Top-5 & 98.57 & 97.55 & 96.92 & 96.82 &  \textbf{98.32} & 98.09 \\ \hline
		\end{tabular}
	\end{center}
	
	\label{svhnblur}
\end{table*}

\paragraph{Mixed Adverse Conditions}
In real-world applications, multiple types of degradation may appear simultaneously.
To this end, we examine if our algorithms remain effective under a mixture of multiple adverse conditions.  We evaluate two settings: 1) first downsampling HQ images by $\alpha$ = 2 and then adding Gaussian noise with std 25; 2) first downsampling HQ images by $\alpha$ = 4 and then blurring with the Gaussian kernel of std 2. We compare HQ, LQ-$\alpha$, RAP-$\alpha$-non-joint, RAP-$\alpha$ and ARAP-$\alpha$-$\beta$, where $\alpha$ and $\beta$ both only consider the downsampling factor for simplicity. ARAP and RAP seamlessly generalize to the mixed adverse conditions, and obtain the most promising performance in Tables \ref{msragausslr} and \ref{msrablurlr}



\begin{figure}
	\centering
	\includegraphics[width=0.95\linewidth]{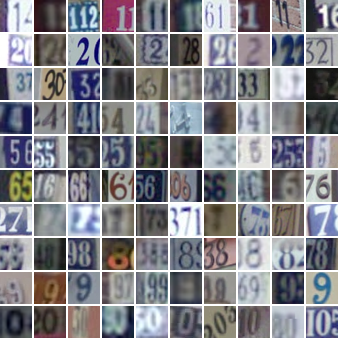}
	\caption{Digit image samples from the SVHN dataset.}
	\label{fig:SVHN}
\end{figure}

\subsubsection{Digit Recognition on the SVHN Dataset}


The Street View House Number (SVHN) dataset \cite{netzer2011reading} contains $73, 257$ digit images of $32\times32$ pixels for training, and $26, 032$ for testing. We focus on investigating the impact of low-resolution and blur on the SVNH digit recognition. Our model has a default configuration of $d = 4$, $d_1 = 2$; $n_1 = 20$, $c_1 = 5$; $n_2 = 50$, $c_2 = 5$; $m_1 = 500$; $m_2 = 10$ (class number used). $conv_1$ is followed by $2\times2$ max pooling.


\paragraph{Low-Resolution}
Table \ref{svhnlr} compares HQ, LQ-$\alpha$, RAP-$\alpha$-non-joint, RAP-$\alpha$ and ARAP-$\alpha$-$\beta$, in the low-resolution case with $\alpha$ = 8. While the LQ-$\alpha$ accuracy drops disastrously, satisfactory top-1 and top-5 accuracy is achieved by ARAP-$\alpha$-$\beta$ ($\beta$ = 16) and RAP-$\alpha$. We observe that more than half of digit images could still be correctly predicted at the extremely low-resolution of $4\times4$ pixels by the proposed methods. 

\begin{table}
	\fontsize{10pt}{12pt}\selectfont
	\caption{The top-1 and top-5 digit recognition accuracy (\%) on the SVHN dataset, where LQ images are downsampling the original images (HQ) by a factor of $\alpha$ = 8.}	
	\begin{center}
		\begin{tabular}{@{\hskip 1mm}c@{\hskip 1mm}|@{\hskip 1mm}c@{\hskip 1mm}|@{\hskip 1mm}c@{\hskip 1mm}|@{\hskip 1mm}c@{\hskip 1mm}|@{\hskip 1mm}c@{\hskip 1mm}|@{\hskip 1mm}c@{\hskip 1mm}}
			\hline
			& HQ & LQ-8 & RAP-8-non-joint & RAP-8 & ARAP-8-16 \\ \hline
			\hline
			Top-1 & 89.23 & 19.60 & 45.98 & 51.00 & \textbf{51.17} \\ 
			Top-5 & 98.57 & 65.44 & 87.08 & \textbf{89.15} & 89.06 \\ \hline
		\end{tabular}
	\end{center}
	
	\label{svhnlr}
\end{table}

\paragraph{Blur}
Table \ref{svhnblur} compares those methods in the Gaussian blur case with standard deviation $\alpha$ = 2. To our astonishments, ARAP-$\alpha$-$\beta$ not only improves over LQ-$\alpha$, but also surpasses the performance of HQ in terms of top-1 accuracy. That is because the original SVNH images (treated as HQ) are real-world photos that unavoidably suffer from certain blur, which can be found in Figure \ref{fig:SVHN}. Convolved with the synthetic Gaussian blur kernel ($\alpha$ = 2), the actual blur kernel's standard deviation becomes larger than 2. 
Hence ARAP-$\alpha$-$\beta$ is potentially able to remove the inherent blurs in HQ images, besides the synthetically added blurs.


\begin{table*}[htb]
	\fontsize{10pt}{12pt}\selectfont
	\caption{The top-1 and top-5 classification accuracy (\%) on the ImageNet validation set, where LQ images are downsampled by a factor of $\alpha$ = 4 or 8.}
	\begin{center}
		\begin{tabular}{c|c|c|c|c|c|c|c}
			\hline
			& HQ & LQ-4 & RAP-4-non-joint & RAP-4 & LQ-8 & RAP-8-non-joint & RAP-8  \\ 
			\hline
			\hline
			Top-1  & 71.46 & 61.92 & 61.16 & 62.03 & 46.67 & 45.37 & 47.22  \\
			Top-5  & 90.62 & 84.13 & 83.65 & 84.35 & 71.55 & 70.60 & 72.32  \\ 
			\hline
		\end{tabular}
	\end{center}
	
	\label{tab:imagenetlr}
\end{table*}

\subsubsection{Image Classification on the ImageNet Dataset}
\label{sec:imagenet}

We validate our algorithm on a large-scale dataset, ImageNet dataset \cite{imagenet}, for image classification of 1,000 classes.
We utilize 1.2 million images of ILSVRC2012 training set for training, and 50,000 images of its validation set for testing.
We study the degradation of low-resolution on the ImageNet image classification.
In our experiment, we customize a popular classification model: \textit{VGG-16} \cite{simonyan2014very} to work on color images directly. 
Specifically, we add three convolutional layers to the beginning of VGG-16, in order to increase the model capacity for handling the low-resolution degradation.
We choose $k = 3, k_p=3$ for $\M_s$ and the configuration of the first three convolutional layers is $n_1$ = 64, $c_1$ = 9; $n_2$ = 32, $c_2$ = 1; $n_3$ = 3, $c_3$ = 5.
The rest architecture is the same as VGG-16.
We use the VGG-16 model released by its authors as the initialization of it, in order to boost the convergence rate.
We follow the conventional protocols in \cite{simonyan2014very} for data pre-processing, including image resizing, random cropping and mean removal of each color channel.

Table \ref{tab:imagenetlr} compares HQ, LQ-$\alpha$, RAP-$\alpha$-non-joint and RAP-$\alpha$, in the low-resolution case with $\alpha$ = 4 and 8.
RAP-4 outperforms LQ-4 and RAP-4-non-joint in terms of both top-1 and top-5 accuracy.
When the low-resolution degradation becomes severe, RAP-8 is superior to LQ-8 and RAP-8-non-joint by a larger margin. 
Specifically, RAP-8 beats LQ-8 by 0.55\% in top-1 accuracy and 0.77\% in top-5 accuracy, and beats RAP-8-non-joint by 1.85\% in top-1 accuracy and 1.72\% in top-5 accuracy, respectively.

\subsubsection{Face Detection on the FDDB Dataset}

We further generalize our proposed algorithm to the face detection task.
We use the training images of the WIDER Face dataset \cite{yang2016wider} as our training set, which consists of 12,880 images and the annotations of 159,424 faces.
and adopt the Face Detection Data Set and Benchmark (FDDB) \cite{fddbTech} as our test set, which contains the annotations for 5,171 faces in a set of 2,845 images.
We study the degradation of low-resolution for the face detection task.
In our experiment, we customize a popular detection model: \textit{Faster R-CNN} \cite{ren2015faster} to work on color images directly. 
Similar to Section \ref{sec:imagenet}, we add three convolutional layers to the beginning of Faster R-CNN, in order to increase the model capacity for handling the low-resolution degradation.
We choose $k = 3, k_p=3$ for $\M_s$ and the configuration of the first three convolutional layers is $n_1$ = 64, $c_1$ = 9; $n_2$ = 32, $c_2$ = 1; $n_3$ = 3, $c_3$ = 5.
The rest architecture is the same as Faster R-CNN.
We use the VGG-16 model in \cite{simonyan2014very} released by its authors as initialization, in order to accelerate the convergence speed.

\begin{figure}[htbp]
	\centering
	\begin{minipage}{0.23\textwidth}
		\centering \subfigure[] {
			\includegraphics[width=\textwidth]{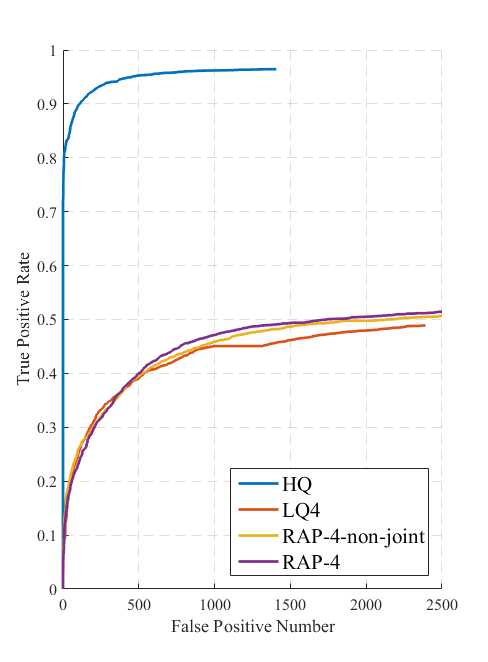}
	}\end{minipage}
	\begin{minipage}{0.23\textwidth}
		\centering \subfigure[] {
			\includegraphics[width=\textwidth]{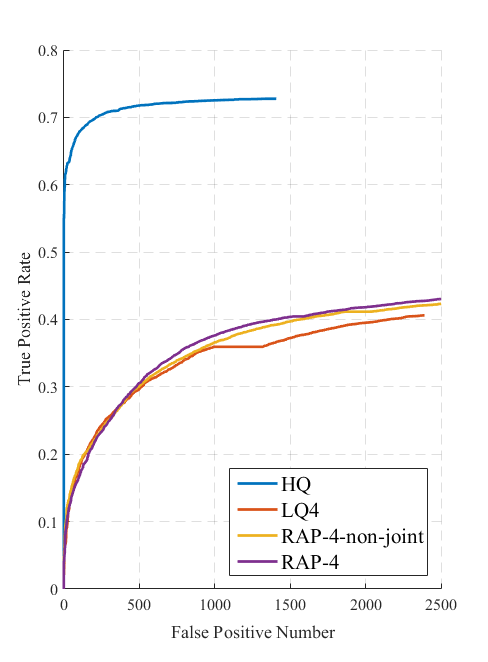}
	}\end{minipage}
	\caption{(a) Discrete ROC curve and (b) Continuous ROC curve on FDDB dataset, where LQ images are downsampled by a factor of $\alpha$ = 4.}
	\label{fig:face_detection}
\end{figure}

Figure \ref{fig:face_detection} shows the discrete and continuous ROC curves of HQ, LQ-$\alpha$, RAP-$\alpha$-non-joint and RAP-$\alpha$, in the low-resolution case with $\alpha$ = 4.
We can observe that there is an obvious performance drop due to the low-resolution degradation.
RAP-4 outperforms LQ-4 and RAP-4-non-joint in terms of recall rate with the same number of false positives.
For example, RAP-4 recalls 50.49\% faces with 2,000 false positives, which is 0.73\% higher than RAP-4-non-joint and 2.55\% higher than LQ-4, respectively.
We obtain the same comparison result in the case of 1,500 false positives, where RAP-4 recalls 48.68\% faces, being 0.67\% higher than RAP-4-non-joint and 3.15\% higher than LQ-4, respectively.

\subsection{Analysis and Visualization}

\subsubsection{Cause of Failure of Deep Models When Using LQ Data}

The failures caused by LQ inputs to deep models can vary a lot, both by degradation type and by the subsequent high-level vision tasks, and sometimes being model-specific too. We pick several examples below to explain this complicacy (all below refer to ordinary deep models without robust pre-training, and our proposed approach alleviates those problems in general):

\begin{itemize}
	\item Low-resolution recognition: the learned filter responses tend to be weak (lack of sharp, recognizable patterns), and inter-channel feature maps tend to be less variable, resulting in poorer discriminativeness of features.
	\item Recognition with salt \& pepper noise: the learned feature maps contain moderately perturbed patterns, and those perturbed patterns can dominate over true features when the noise level gets high (e.g., when 50\% pixels are corrupted) and thus confuse classification. 
	\item Low-resolution object detection: when the two-stage detectors are used, low-resolution causes fewer object proposals to be generated in the first region proposal network stage. As a proof-of-concept, we tried to generate proposals on high-resolution images, then applied those bounding box locations to low-resolution images to train classifiers, finding the performance loss to be restored to a large extent. The same observations were found in hazy images~\cite{li2017aod}. 
\end{itemize}

\subsubsection{Convolutional and Additive Adverse Conditions}
We have tested four adverse conditions so far. RAP and ARAP improves the recognition in all cases, which shows that the pre-training of image restoration achieves feature enhancement in the recognition model and benefits the visual recognition task.  
We note that low-resolution and blur clearly receive extra bonus from ARAP than RAP. In the other two cases, i.e., noise and occlusion, RAP and ARAP perform approximately the same. Such contrastive behaviors hint that some adverse conditions might be more suitable for ARAP to perform than the others. 

In the general image degradation model, the observed image $Y$ is usually represented as
\begin{equation}
Y = \mathcal{F} \ast X + e,
\end{equation}
where $\mathcal{F}$ denotes the point spread function, $X$ is the clean image, and $e$ is the noise. Low-resolution and blur are usually modeled in $ \mathcal{F}$ as low-pass filters, while noise and occlusion can be incorporated in $e$ as additive perturbations. We term the former category as \textbf{convolutional adverse conditions}, and the latter as \textbf{additive adverse conditions}. 
We conjecture that the additive adverse condition causes pixel-wise corruptions but still retains some structural information, 
while the convolutional adverse condition results in global detail loss and smoothening, which may be more challenging for recognition and thus needs more robust feature extractions by purposely pre-training $\M_s$ in heavier adverse conditions.
This hypothesis will be further justified experimentally when we extend our framework to video cases.



\subsubsection{Effects of End-to-End Tuning in RAP}
\begin{figure}[htbp]
	\centering
	\begin{minipage}{0.46\textwidth}
		\centering {
			\includegraphics[width=\textwidth]{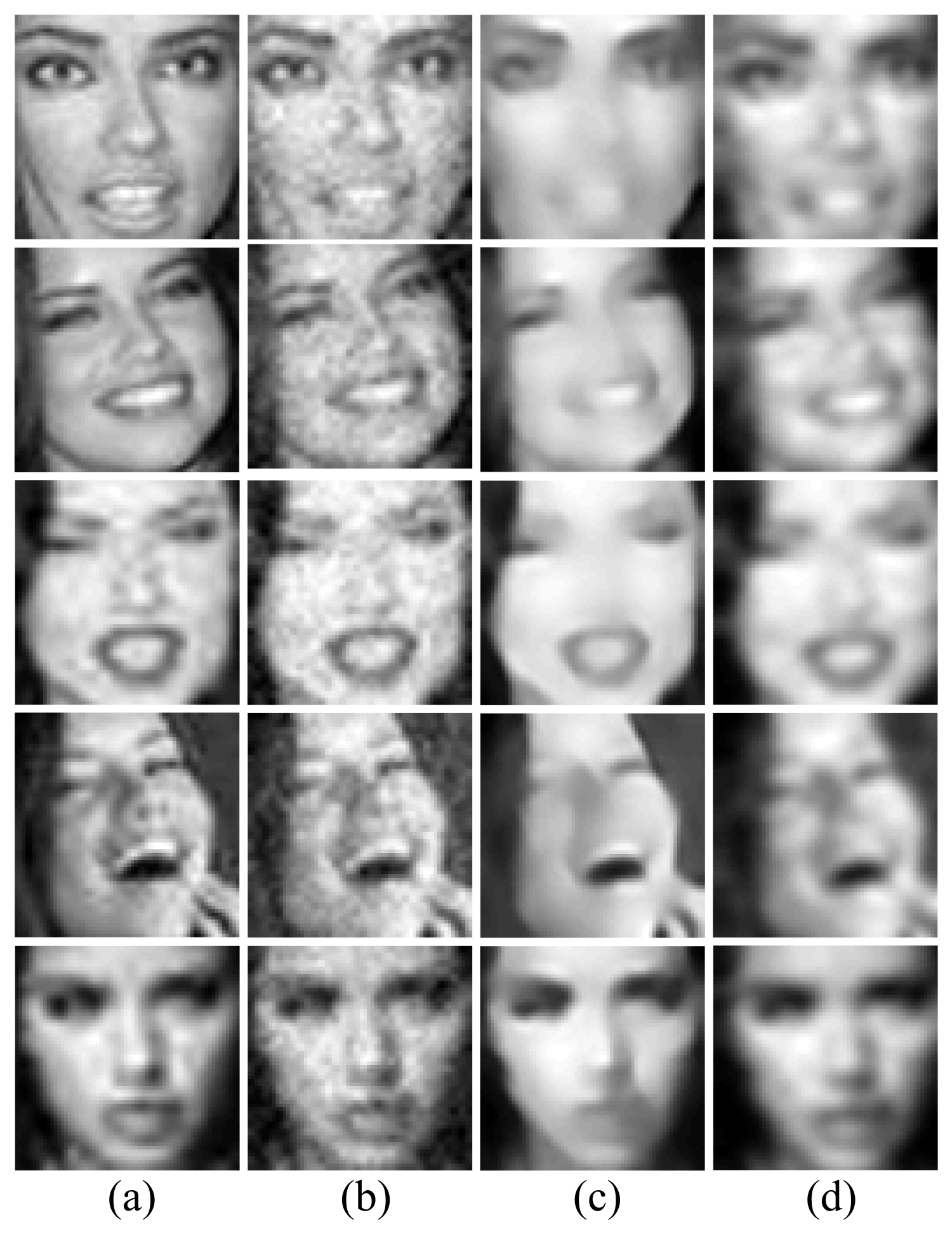}
	}\end{minipage}
	\caption{Visualized features for successful examples of joint tuning, i.e. those correctly classified by RAP but misclassified by RAP-non-joint. Column (a): original HQ images from MSRA-CFW. (b): LQ images from the first mixed adverse condition setting. (c): visualized $\mathcal{F}_k$ (intermediate features by RAP-non-joint). (d): visualized $\mathcal{F}'_k$ (intermediate features by RAP). }
	\label{success}
\end{figure}

To further analyze our proposed RAP, we focus on the following two questions:
How the joint tuning of $\M$ modifies the features learned in the pre-trained $\M_s$, and why it improves the recognition in almost all adverse conditions? 

To answer these questions, we visualize and compare the features in the first $k_p$-th layers of $\M$ before and after the end-to-end tuning, denoted as $\mathcal{F}_k$ and $\mathcal{F}'_k$, respectively.
Recall that in the pre-training step of RAP, $\M_s$ reconstructs the images by feeding $\mathcal{F}_k$ to $k-k_p$ additional layers, that are removed in the joint tuning step. We pass both $\mathcal{F}_k$ and $\mathcal{F}'_k$ through the fixed mapping of these $k-k_p$ layers (obtained when training $\M_s$). The output, which is of the same dimension as HQ images, is used to visualize of $\mathcal{F}_k$ or $\mathcal{F}'_k$. Note that the visualizations of $\mathcal{F}_k$ are just the reconstruction results of $\M_s$. 

Figure \ref{success} presents feature visualizations for five MSRA-CFW images that are correctly classified by RAP but misclassified by RAP-non-joint. As shown in column (c), the $\mathcal{F}_k$ features from the un-tuned $\M_s$ are heavily over-smoothed, with much discriminative information lost. In contrast, the visualizations of $\mathcal{F}'_k$ yield a few impressive restoration results in column (d). The joint tuning step enables the closed-loop consideration of two information sources (HQ data and labels) for two related tasks (restoration and recognition).  It thus boosts not only the recognition accuracy, but also the restoration: column (d) results contain much richer and finer details, and are apparently more recognizable than column (c).

\begin{figure*}[htbp]
	\centering
			\includegraphics[width=0.9\textwidth]{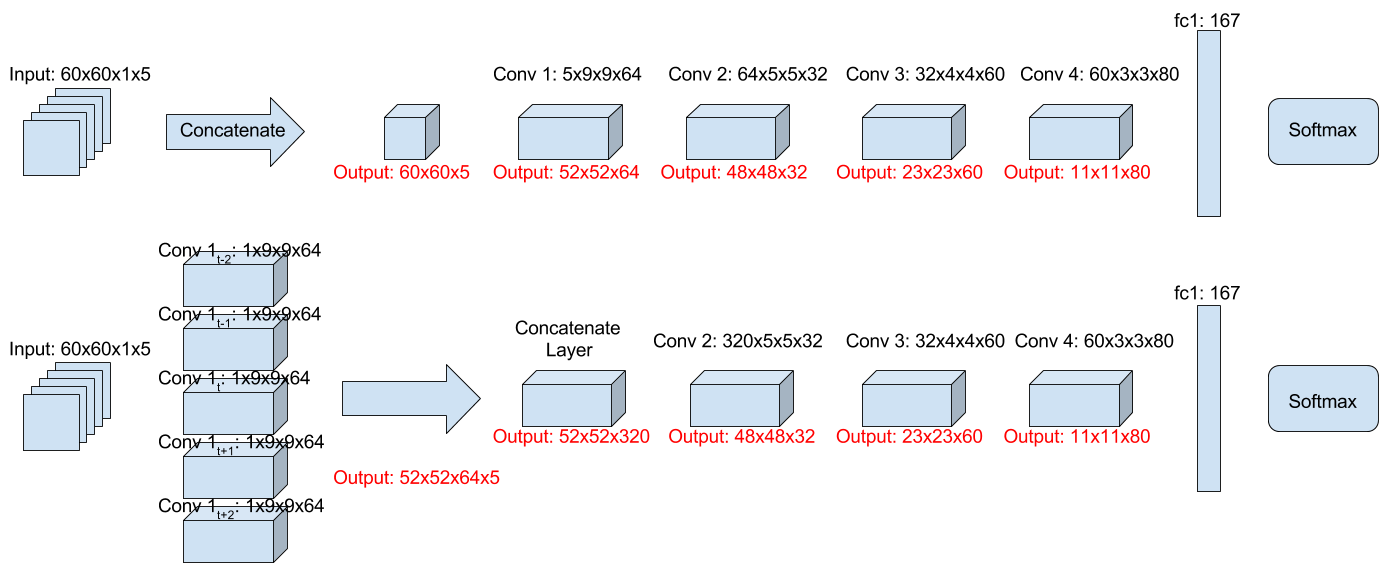}
	\caption{Model architectures for YTF video recognition experiments. Top: early fusion. Bottom: slow fusion.}
	\label{fig:video_net}
\end{figure*}

\section{Video Recognition in Adverse Conditions}
\label{sec:video}

\subsection{Temporal Fusion for Video Based Models}

Temporal fusion of feature representations is usually adopted in deep learning based methods for video-related tasks.
Karpathy et al.\ \cite{karpathy2014large} first provided an extensive empirical evaluation of CNNs on large-scale video classification. 
In addition to the single frame baseline, \cite{karpathy2014large} discussed three connectivity
patterns. The \textbf{early fusion} combines frames within a time window immediately in the
pixel level. 
The \textbf{late fusion} separately extracts features from each frame  and does not merge them until the first fully connected layer.
The \textbf{slow fusion} is a balanced mix between the two, which slowly unifies temporal information throughout the network by progressively merging features from individual frames. 

\subsection{Robust Adverse Pre-training for Video Recognition}

Following \cite{karpathy2014large,kappeler2016video,liu2018learning}, we treat each video as a number of short, fixed-sized clips. Each clip is set to contain $2T + 1$ contiguous frames in time.
The video based CNN model $\M_v$ takes a clip as its input. To extend $\M_v$ to adverse conditions, we first pre-train a single image model $\M$ using RAP or ARAP, by treating all frames as individual images and formulating an image based recognition problem. We then convert $\M$ to $\M_v$ based on different fusion strategies, and initialize the weights of $\M_v$ from $\M$ using the weight transfer proposed in \cite{kappeler2016video}. $\M_v$ is then tuned in the video setting. Since we find the late fusion results to be always inferior to the other two, we omit discussing the case of late fusion hereinafter.

For early fusion, we copy the $conv_1$ layer of $\M$ ($n_1$ filters of $c_1 \times c_1$) for $2T + 1$ times, and divide the weights of all filters by $2T+1$\footnote{
Detailed reasoning follows Section III.C of \cite{kappeler2016video}. Our early and slow fusion models resemble their architectures (a) and (b). }. 
We then use them in the new $conv_1$ layer of $\M_v$ with the size $n_1 \times c_1 \times c_1 \times (2T+1)$, to fuse information in the first layer.
All other layers of $\M_v$ are identical with $\M$ in both configuration and weight transfer. 

For slow fusion, we copy the $conv_1$ layer of $\M$ for $2T + 1$ times into the new $conv_1$ layer of $n_1 \times c_1 \times c_1 \times (2T+1)$, without changing the weights. 
We then stack the filters of the $conv_2$ layer of $\M$  ($n_2$ filters of $c_c \times c_c$)  for $2T + 1$ times and divide all weights by $2T+1$, constituting the new $conv_2$ layer of $n_2 \times c_2 \times c_2 \times (2T+1)$ to fuse information in the second layer.
All other layers of $\M_v$ remain identical to $\M$.

%

\subsection{Experiments on Benchmarks}

We use a video face dataset: the \textit{YouTube Face} (YTF) benchmark \cite{wolf2011face} to validate our algorithm.
We choose the 167 subject classes that contain 4 video sequences.
For each class, we randomly pick one video for testing and the rest for training. The face regions are cropped using the given bounding boxes. As the majority of cropped faces have side lengths between 56 and 68, we slightly resize them all to $60 \times 60$ for simplicity, and refer to those as the \textit{original YTF set} hereinafter. We densely sample clips of 5 $(T = 2)$ frames  from each video with a stride of one frame, and present each clip individually to the model. The class predictions are averaged to produce an estimate of the video-level class probabilities. For the single image model,  we chose $d = 5$, $d_1 = 4$, with each layer: $n_1$ = 64, $c_1$ = 9; $n_2$ = 32, $c_2$ = 5; $n_3$ = 60, $c_3$ = 4; $n_4$ = 80, $c_4$ = 3, $m_1$ = 167. All video based models start from the same pre-trained single frame model, and then split filters differently. We enforce filter symmetry as in \cite{kappeler2016video}.  The detailed architectures are drawn in Figure~\ref{fig:video_net}.

\begin{table*}[t]
	\fontsize{10pt}{12pt}\selectfont
	\caption{The top-1 and top-5 accuracy (\%) on YTF, in the low resolution setting, with different fusion strategies.}
	\begin{center}
		\begin{tabular}{c|c|c|c|c|c|c}
			\hline
			& & HQ & LQ-2 & RAP-2 & ARAP-2-4 & ARAP-2-8 \\ \hline
			\hline
			Single  & Top-1 & 37.32 & 38.30 & 39.16 & 41.05 & 38.58 \\
			Frame & Top-5 & 60.01 & 59.56 & 59.94 & 61.97 & 60.33 \\ 
			\hline
			\hline
			Early  & Top-1 & 38.11 & 37.73 & 39.83 & \textbf{41.11} & 38.05 \\
			Fusion & Top-5 & 58.48 & 62.42 & 62.74 & \textbf{63.85} & 60.79 \\ 
			\hline
			\hline
			Slow  & Top-1 & 35.99 & 37.76 & 39.60 & 40.98 & 39.67\\
			Fusion & Top-5 & 53.20 & 58.79 & 60.86 & 63.03 & 61.50 \\ 
			\hline
		\end{tabular}
	\end{center}
	
	\label{YTF_lr}
\end{table*}

Similarly to image based experiments, Tables \ref{YTF_lr} and \ref{YTF_noise} compare HQ, LQ-$\alpha$, RAP-$\alpha$, and ARAP-$\alpha$-$\beta$, in the settings of low resolution ($\alpha$ = 2) and salt \& pepper noise ($\alpha$ = 50\%). ARAP/RAP bring substantially improved performance within each fusion. Recall that the best fusion models in \cite{karpathy2014large} displayed only modest improvement over single frame models (from 59.3\% to 60.9\%), we consider that our 1.11\% top-5 gain by early fusion in the low resolution setting, and 13.37\% top-5 gain by slow fusion in the noise setting, are both reasonably good. 

While \cite{karpathy2014large} advocated slow fusion for normal visual recognition problems, the situations seem more complicated when adverse conditions step in. Our results imply that additive adverse conditions favor slow fusion, while convolutional adverse conditions prefer early fusion. 
We tried experiments in the blur case, whose observations are close to the low resolution case. 
We conjecture that  early fusion becomes the preferred option when the data is already heavily ``filtered'' by degradation operators or blur kernels, such that it cannot afford extra information loss after more filtering. The diverse fusion preferences manifest the unique complication brought by adverse conditions. 

\begin{table}[tbp]
	\fontsize{10pt}{12pt}\selectfont
	\caption{The top-1 and top-5 accuracy (\%) on YTF, in the salt \& pepper noise setting, with different fusion strategies.}
\begin{center}
\begin{tabular}{c|c|c|c|c}
\hline
 & & HQ & LQ-50\% & RAP-50\% \\ \hline
 \hline
Single  & Top-1 & 37.32 & 15.81 & 31.64 \\
 Frame & Top-5 & 60.01 & 30.93 & 48.48 \\ 
\hline
\hline
Early  & Top-1 & 38.11 & 18.86 & 21.20  \\
Fusion & Top-5 & 58.48 & 36.59 & 38.01  \\ 
\hline
\hline
Slow  & Top-1 & 35.99 & 21.97 & \textbf{34.55} \\
Fusion & Top-5 & 53.20 & 39.00 & \textbf{52.37}  \\ 
\hline
\end{tabular}
\end{center}

\label{YTF_noise}
\end{table}

As the last finding, in the low resolution case, the RAP and ARAP results using LQ data can even surpass HQ results notably. We input the original YTF set to the trained ARAP-2-4 models, and also witnesses much improved accuracy in Table \ref{YTF_HQ}, than feeding the same set through the HQ models. The best top-1 and top-5 results in Table \ref{YTF_HQ} also surpass all results in Table \ref{YTF_lr}. We suspect that although the original YTF set is treated as clean and high-quality, it was actually contaminated by degradations during image collection, and is thus low-quality from that viewpoint. Applying RAP and ARAP compensates  part of the unknown information loss. From another perspective, training a model on LQ data and then applying on HQ data is related to a special  data augmentation introduced in \cite{vlrr}, that blends HQ and LQ data for training. While \cite{vlrr} confirmed its effectiveness in recognizing LQ subjects, we discover its usefulness for normal (HQ) visual recognition too.


\begin{table}[h]
	\fontsize{10pt}{12pt}\selectfont
	\caption{The top-1 and top-5 accuracy (\%) by feeding the original YTF set to the trained ARAP-2-4 models.}
\begin{center}
\begin{tabular}{c|c|c|c}
\hline
 & Single Frame & Early Fusion & Slow Fusion \\ \hline
 \hline
Top-1 & 41.31 & 41.60  & \textbf{42.20} \\ 
Top-5 & 62.30 & \textbf{64.04} & 63.10\\ \hline
\end{tabular}
\end{center}

\label{YTF_HQ}
\end{table}



\section{Coping with Unknown Adverse Conditions: A Transfer Learning Approach}
\label{sec:unknown}

In all previous experiments, we train with $\{\mathbf{x}_i, \mathbf{y}_i\}_{i=1}^N$ pairs. That is equivalent to assuming a pre-known degradation process from $\{\mathbf{x}_i\}_{i=1}^N$ to $\{\mathbf{y}_i\}_{i=1}^N$. Such an assumption, as made in \cite{vlrr}, is impractical for real-world LQ data and restricts our experiments to synthesized test data so far. In this section, we develop a transfer learning approach to significantly relax this strong assumption. It ensures the wide applicability of our algorithms, even when the degradation parameters cannot be accurately inferred.

For convolutional adverse conditions, the recognition accuracy is usually peaked at some optimal $\beta^* > \alpha$.
The additive adverse conditions seems insensitive to $\beta$. However, the performance ARAP-$\alpha$-$\beta$ results ($\beta > \alpha$) are observed to be always better, or at least comparable to ARAP-$\alpha$, even when $\beta$ deviates far away from $\beta^*$.

\begin{algorithm}[ht]
	\caption{Transfer ARAP Learning}
	\label{transferA}
	\begin{algorithmic}[1]
		\REQUIRE Configurations of $\M$ and $\M'$; the choice of $k$; the clean source dataset $\{\mathbf{x}_i\}$ and $\{l_i\}$; the target dataset $\{\mathbf{x}'_i\}$ and $\{l'_i\}$, with unknown $\alpha'$.
		
		\STATE Decide the major degradation type in $\{\mathbf{x}'_j\}$, and choose $\beta'$ such that it overestimates $\alpha'$.
		
		\STATE Generate $\{\mathbf{y}_i\}$ from $\{\mathbf{x}_i\}$, based on the degradation processes of the major type, parameterized by $\beta'$.
		
		\STATE Perform Steps 3 - 6 in Algorithm 1, to train $\M$ on the source dataset. 
		
		\STATE Export the first $k$ layers from $\M$ to initialize the first $k$ layers of $\M'$.
		
		\STATE Tune $\M'$ over \{$\{\mathbf{x}'_i\}$, $\{l'_i$\}\}.
		
		\ENSURE $\M'$.
	\end{algorithmic}
\end{algorithm}

On a target dataset with real-world corruptions, it is reasonable to assume that the major type of adverse condition(s) can still be identified, but the parameter $\alpha'$ of the underlying degradation process cannot be accurately estimated. Observing the robustness of ARAP w.r.t. $\beta$, we propose the \textit{Transfer ARAP Learning} (\textbf{T-ARAP}) approach, as detailed in Algorithm \ref{transferA}. The core idea is to first choose $\beta'$ that we empirically believe $\beta' > \alpha'$, then performing RAP (with $\beta' $) to train $\M$ on a source dataset. Next, we transfer the learned sub-model of $\M$ to initialize $\M'$, which is later tuned for the target dataset. Note that $\beta'$ is not necessarily very close to $\alpha'$. In practice, one may safely start with some large $\beta'$, and scan backwards for an optimal value.  

\begin{table}[t]
	\fontsize{10pt}{12pt}\selectfont
	\caption{The top-1 and top-5 accuracy (\%) on the original YTF set, by transferring from the MSRA-CFW RAP-4 model.}
	\begin{center}
		\begin{tabular}{c|c|c|c|c}
			\hline
			& LQ$_d$ & LQ$_p$ & T-ARAP-non-joint & T-ARAP\\ \hline
			\hline
			Top-1 & 32.65 & 32.35 & 33.67 & \textbf{34.77} \\ 
			Top-5 & 45.37 & 47.73 & 48.11 & \textbf{53.11}\\ \hline
		\end{tabular}
	\end{center}
	
	\label{transfer}
\end{table}

We validate the approach via conducting the following experiment: improving face identification on (original) YTF by referring to a RAP model on MSRA-CFW. For simplicity, here we perform the task of single-image face identification, and treat the original YTF set as an image collection without utilizing temporal coherence. We visually observe that the original YTF images have inherently lower quality, which is also supported by Table \ref{YTF_HQ}. We select low resolution as our target adverse condition, and not too aggressively, choose $\beta'$ = 4. We hence take the $\M_s$ part from the RAP-4 model trained on MSRA-CFW, to initialize the first 2 layers of $\M'$. Meanwhile, we design three baselines for comparison: 
1) \textbf{LQ}$_d$ model trained directly end-to-end on YTF; 
2) \textbf{LQ}$_p$ model trained on YTF with classical unsupervised layer-wise pre-training; 
3) \textbf{T-ARAP-non-joint}, taking the untuned $\M_s$ of RAP-4 for $\M'$ initialization. In Table \ref{transfer}, T-ARAP improves the top-5 recognition accuracy by nearly 8\% over the naive \textbf{LQ}$_d$, with no strong prior knowledge about the degradation process nor its parameter, 
which demonstrates the effectiveness of our proposed transfer learning approach.



%
%
%


\section{Conclusion and Discussions}
\label{sec:conc}

This paper systematically improves deep learning models via robust pre-training for image and video recognition under adverse conditions. We thoroughly evaluate our proposed algorithm on various datasets and degradation settings, and analyze our results in depth, 
which shows the effectiveness of our proposed algorithm.
A transfer learning approach is also proposed to enhance the real-world applicability.  

\bibliographystyle{IEEEtran}
\bibliography{vlqr_tip}

%

\begin{IEEEbiography}[{\includegraphics[width=1in,height=1.25in,clip,keepaspectratio]{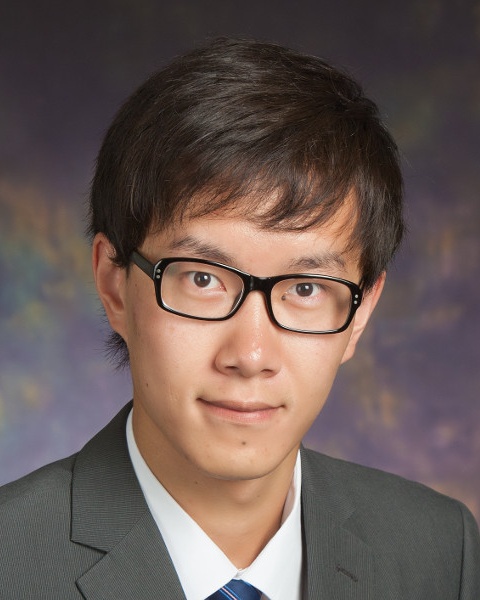}}]{Ding Liu}
(S'15--M'18) received the B.S. degree from the Chinese University of Hong Kong, Hong Kong, in 2012, and the M.S. and Ph.D. degrees from the University of Illinois at Urbana-Champaign, USA, in 2014 and 2018, respectively. 
His research expertise encompasses image restoration and image enhancement. He has research interests in the broad area of computer vision, image processing and deep learning.
\end{IEEEbiography}

\begin{IEEEbiography}[{\includegraphics[width=1in,height=1.25in,clip,keepaspectratio]{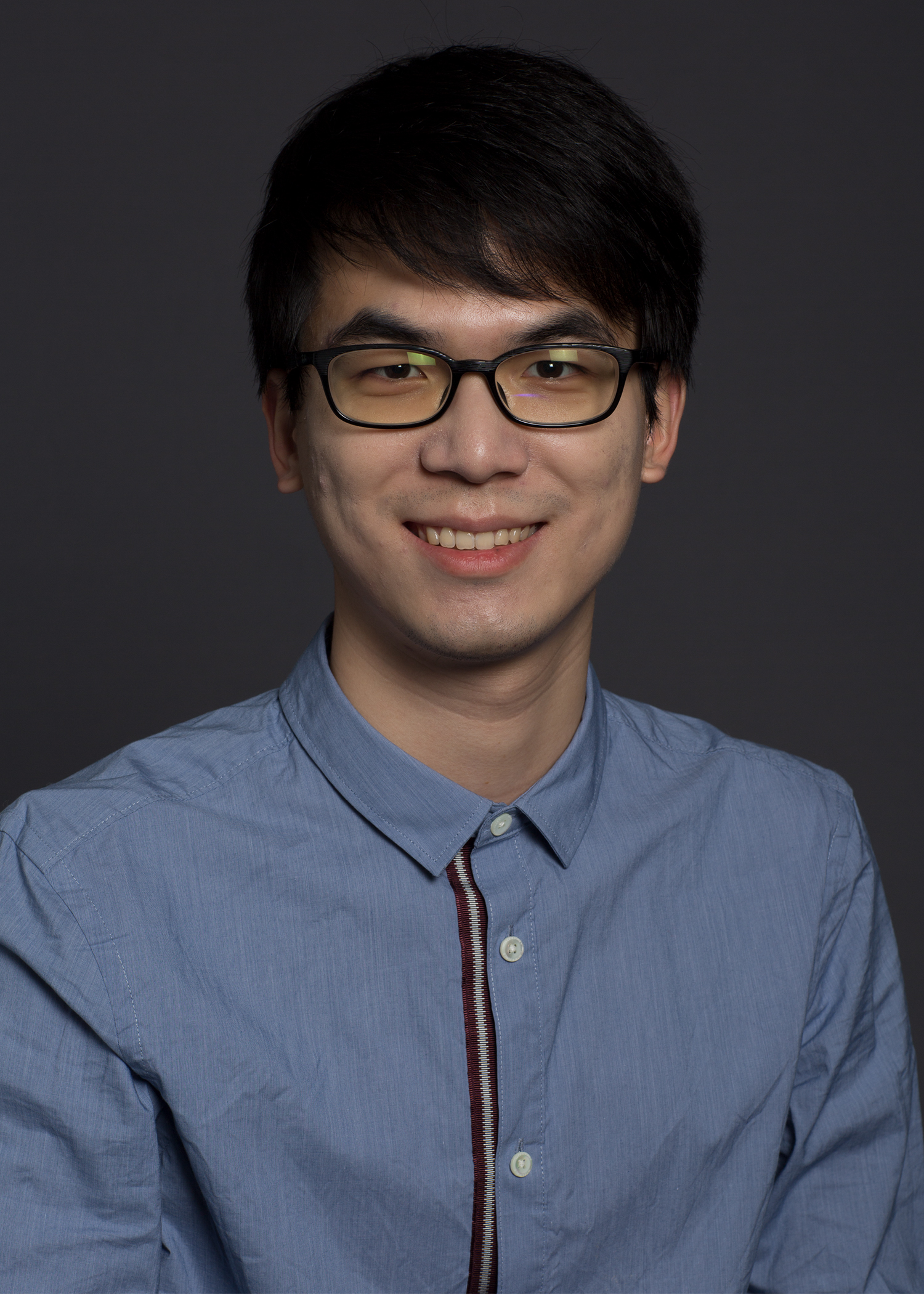}}]{Bowen Cheng}
is a Ph.D. student in Electrical and Computer Engineering at University of Illinois at Urbana-Champaign (UIUC). His Ph.D. advisor is Prof. Thomas Huang and he is doing research in computer vision. Specifically, he works on object detection, image classification and semantic segmentation. Before commencing his graduate studies, he received his B.S. in Electrical and Computer Engineering at UIUC in 2017.
\end{IEEEbiography}

\begin{IEEEbiography}[{\includegraphics[width=1in,height=1.25in,clip,keepaspectratio]{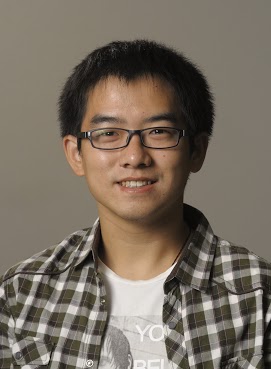}}]{Zhangyang Wang}
is an Assistant Professor of Computer Science and Engineering (CSE), at the Texas A\&M University (TAMU). During 2012-2016, he was a Ph.D. student in the Electrical and Computer Engineering (ECE) Department, at the University of Illinois at Urbana-Champaign (UIUC), working with Professor Thomas S. Huang. Prior to that, he obtained the B.E. degree at the University of Science and Technology of China (USTC), in 2012. He was a former research intern with Microsoft Research (summer 2015), Adobe Research (summer 2014), and US Army Research Lab (summer 2013). Dr. Wang’s research has been addressing machine learning, computer vision and multimedia signal processing problems, as well as their interdisciplinary applications, using advanced feature learning and optimization techniques. He has co-authored over 70 papers, and published several books and chapters. He has been granted 3 patents, and has received around 20 research awards and scholarships. 
\end{IEEEbiography}

\begin{IEEEbiography}[{\includegraphics[width=1in,height=1.25in,clip,keepaspectratio]{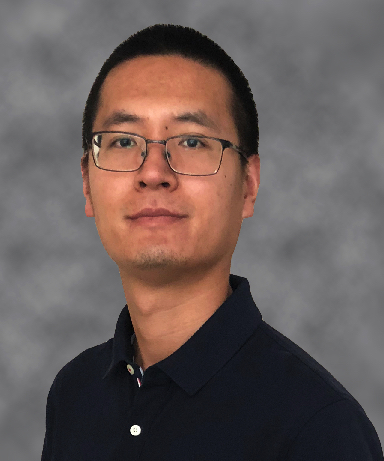}}]{Haichao Zhang}
received the Ph.D. degree in Computer Science from Northwestern Polytechnical University, Xi’an, China.  He was with the Beckman Institute for Advanced Science and Technology, University of Illinois at Urbana-Champaign as a visiting scholar from 2009 to 2011 and the Rhodes Information Initiative at Duke University as a Post-Doctoral Associate from 2013 to 2014. From 2014 to 2016, he worked as a Research Scientist at Amazon Go, Seattle, WA. Since 2016, he has been a Senior Research Scientist with Baidu Research, Sunnyvale, CA.
\end{IEEEbiography}

\begin{IEEEbiography}[{\includegraphics[width=1in,height=1.25in,clip,keepaspectratio]{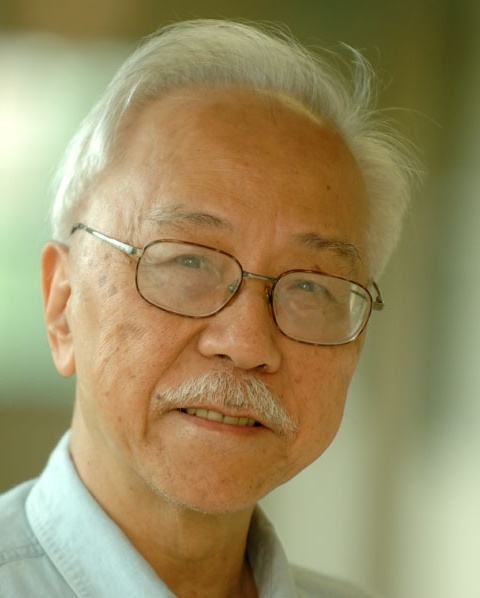}}]{Thomas S. Huang}
(F'01) received the B.S. degree in electrical engineering from National Taiwan University, Taipei, Taiwan, R.O.C., and the M.S. and Sc.D. degrees in electrical engineering from the Massachusetts Institute of Technology (MIT), Cambridge. He was on the Faculty of the Department of Electrical Engineering at MIT from 1963 to 1973; and on the Faculty of the School of Electrical Engineering and Director of its Laboratory for Information and Signal Processing at Purdue University from 1973 to 1980. In 1980, he joined the University of Illinois at Urbana-Champaign, where he is now William L. Everitt Distinguished Professor of Electrical and Computer Engineering, and Research Professor at the Coordinated Science Laboratory, and at the Beckman Institute for Advanced Science he is Technology and Co-Chair of the Institute’s major research theme Human Computer Intelligent Interaction. His professional interests lie in the broad area of information technology, especially the transmission and processing of multidimensional signals. He has published 21 books, and over 600 papers in network theory, digital filtering, image processing, and computer vision.
Dr. Huang is a Member of the National Academy of Engineering; a Member of the Academia Sinica, Republic of China; a Foreign Member of the Chinese Academies of Engineering and Sciences; and a Fellow of the International Association of Pattern Recognition, IEEE, and the Optical Society of America. Among his many honors and awards: Honda Lifetime Achievement Award, IEEE Jack Kilby Signal Processing Medal, and King-Sun Fu Prize of the Int. Asso. for Pattern Recognition.
\end{IEEEbiography}








\end{document}